\documentclass{article}
\usepackage{spconf,amsmath,amssymb,graphicx,booktabs}
\usepackage[table]{xcolor}
\definecolor{fuse}{RGB}{226,236,247}
\usepackage[hidelinks]{hyperref}
\usepackage{url}

\newcommand{\phasefusion}{\textsc{PhaseFusion}}
\newcommand{\spr}{\textsc{SPR}}
\makeatletter\def\paragraph#1{\par\medskip\noindent\textbf{#1}\ }\makeatother

\title{Stable Scores, Unstable Answers: Frame Phase and Option Order\\
in Video Multiple-Choice Evaluation}

\name{Lichen Zhu, Yiheng Wang, Yueqian Lin, Hai ``Helen'' Li, Yiran Chen}
\address{Duke University, Durham, NC, USA}

\begin{document}
\ninept
\maketitle

\begin{abstract}
Video-language models are ranked by multiple-choice accuracy on frames from a uniform grid. The grid has two parameters, a rate and a phase,
and benchmarks report only the rate. The phase moves answers: two deployed samplers differing only by a half-step phase offset answer 23.6\% of
questions differently while scoring within a point, and across four releases from two families shifting only the phase changes roughly one answer in five
after controlling option order. \phasefusion{} decodes three offset grids and averages the option posteriors. The grids are the polyphase components of the dense grid. Fusion matches a 32-frame single pass in accuracy within a prespecified margin (logit-scored) and cuts the answers a half-step shift of all three grids changes from 18.2\% to 10.1\%. Option order, which changes only the presentation, is flagged instead by a one-pass answer margin. Report the phase convention with the budget, or marginalize it.
\end{abstract}


\begin{figure*}[t]
\centering
\includegraphics[width=\textwidth]{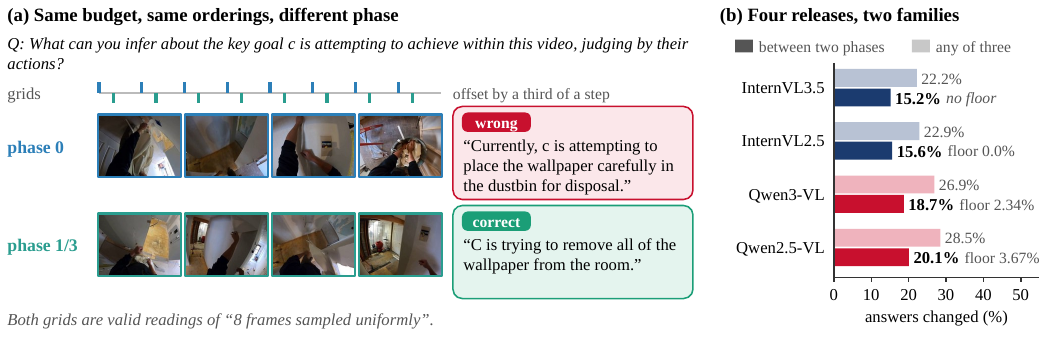}
\caption{\textbf{The sampling phase changes answers, not scores.} (a) One EgoSchema clip decoded at two phases of the \emph{same} uniform 8-frame grid, whose sample times are drawn on a common axis (four of eight frames shown per phase). Each phase is unanimous across all ten option orderings, yet they pick different options, one of them wrong. (b) Four releases from two families. Dark bars: consensus answers that differ between two phases, with each family's measured re-run floor printed beside the bar (InternVL3.5 has no re-run corpus). Light bars: answers that change under any
of the three phases.}
\label{fig:phase}
\end{figure*}

\section{Introduction}
\label{sec:intro}

Video-language leaderboards rest on a common front end: a clip of $T$ frames
is thinned to $b$ frames on a uniform grid, the model answers a multiple-choice
question, and the fraction of answers that match the key is the score
\cite{mangalam2023egoschema, li2024mvbench, fu2025videomme}. The score is
trusted because re-running the harness returns the same number. A uniform grid,
however, has two parameters, and only one of them is ever written down. The
rate, the frame budget $b$, appears in every protocol. The phase, where the
first sample falls, is left to whichever line of code draws the indices, and we
know of no benchmark protocol that states it. A re-run holds that phase fixed
along with everything else, so the reproducibility that earns the score its
trust is measured under the one condition that hides the parameter.

Prior audits of video evaluation have varied \emph{which} and \emph{how many}
frames are fed \cite{buch2022revisiting, brkic2025frame, ok2025tempcore}. Sun
et al.~\cite{sun2026stablecurves} vary the frame budget and find stable curves
over unstable items, and Cao et al.~\cite{cao2026caved} use an offset shift as
an intervention. All of these effects are read off accuracy. A parallel line
documents sensitivity to the presentation of the options themselves
\cite{zheng2024robust, pezeshkpour2024order, robinson2022leveraging,
zong2024fool, liu2024mmbench, facetprobe2026}, including whether the correct
option is present at all \cite{wang2026absent}. Closest in name, PAS
\cite{sun2025pas} averages over phase offsets of the rotary temporal encoding
inside the model. The phase studied here is different in kind: a free parameter
inside one fixed sampling strategy, at an unchanged rate and outside any model,
whose effect is an answer-level reproducibility property rather than an
accuracy delta. We therefore hold the rate, vary the phase, and measure answers
rather than scores.

This paper asks how much that unreported phase matters and what an evaluator
should do about it. We first measure it. Option order is itself a large
nuisance, so we control it before looking at the phase, and
Section~\ref{sec:audit} reports the \emph{consensus answer} over ten option
orderings. Against that baseline, shifting only the phase changes roughly one
answer in five (Fig.~\ref{fig:phase}) while accuracy stays flat, separates two
samplers that are already in deployment, and survives an eightfold increase in
the frame budget. We then remove it rather than legislate it.
Section~\ref{sec:method} averages the model's option posteriors over offset
grids, the classical multi-view test-time operator
\cite{krizhevsky2012imagenet, wang2016tsn, feichtenhofer2019slowfast,
shanmugam2021better}, applied here to an axis no benchmark protocol reports. In
this setting the operator has an exact structure: the $S$ offset grids are the
polyphase components of a dense $Sb$-frame grid, so averaging over them
integrates the phase out while the same frames are decoded in three short
passes. The two nuisances call for different treatments. The phase changes the
evidence, so it can be averaged out. The order changes only its presentation,
and averaging it exhaustively is factorial, so we flag it instead
(Section~\ref{sec:flag}).

\section{Phase Sensitivity of Uniform Frame Sampling}
\label{sec:audit}

\subsection{Protocol}
The core $3{\times}3$ audit evaluates Qwen2.5-VL-7B (Q2.5) \cite{bai2025qwen25vl}, Qwen3-VL-8B (Q3) \cite{qwen2025qwen3vl}, and InternVL2.5-8B (IV2.5) \cite{chen2024internvl25} on EgoSchema \cite{mangalam2023egoschema}, MVBench \cite{li2024mvbench}, and Video-MME \cite{fu2025videomme}: nine 500-question cells with 10 shared orderings. The phase sweep, three passes per ordering, re-decodes a 200-question subset of each cell (593--599 items per model) under the same ten orderings, and ordering results use all 500. In all, the study decodes 289k passes. Separate replications use MiMo-VL-7B-RL \cite{mimo2025vl} (450 logit-scored and 1{,}495 fusion items) and InternVL3.5 \cite{wang2025internvl35}.

Letters map back to a stable \emph{content index} through the inverse
permutation, and a cell's \emph{consensus answer} is the plurality content
index over its ten orderings, ties broken by lowest index.

All decoding is greedy argmax. Within a run it is bit-stable: 2{,}952 duplicated permutations agree 100\%. The re-run floor is measured, not assumed, following the same-ordering
control of Facet-Probe \cite{facetprobe2026}: phase 0 is re-decoded in a
separate run (same weights, frames, prompts, greedy) and compared with the
released corpus at the consensus level. GPU kernel nondeterminism alone puts it at 2.34\% for Qwen3-VL and 3.67\% for Qwen2.5-VL ($n{=}599$ each). InternVL2.5 re-derives its corpus bit-exactly (0.0\% of 5{,}930 passes), and InternVL3.5 enters the phase sweep only.

\noindent\textbf{The ordering nuisance, and why it goes first.}
Per-cell flip rates span 40.8--66.4\% over the nine 500-question cells (pooled 54.0\%, $n{=}4500$). Across the nine cells the reported score moves by at most 5.8 points
under the same re-orderings. In total, 39.2\% of a single
ordering's gold-correct answers are wrong under at least one other ordering
(9.3\% against the ten-ordering consensus outright): a single ordering is a
lottery ticket for two correct answers in five, and on MVBench and Video-MME one
of the ten draws reverses the InternVL2.5 vs.\ Qwen3-VL ranking that the other
nine report.

\subsection{The sampling phase}
Shifting only the phase changes roughly one answer in five while accuracy
stays flat. Between two phases of the same 8-frame grid, under the same ten
orderings, the consensus answer differs on 18.7\% of items for Qwen3-VL and
20.1\% for Qwen2.5-VL, $8.0\times$ and $5.5\times$ their re-run floors, and on
15.6\% and 15.2\% for the two InternVL releases (Fig.~\ref{fig:phase}). Per
benchmark the rate runs from 11.2\% (MVBench, $n{=}199$ items, 95\% CI
7.7--14.9) to 29.5\% (Video-MME, $n{=}200$, 24.5--34.7), both on Qwen2.5-VL,
and clears the floor in every cell. Per-phase accuracy moves by less than two
points.

Allowing all three phases, the answer changes at least once on 26.9\% of
questions for Qwen3-VL and 28.5\% for Qwen2.5-VL, and on 22.9\% and 22.2\% for
the InternVL releases (Fig.~\ref{fig:phase}b). Ties do not drive this. On the
84--86\% of items whose consensus is tie-free the three-phase rate is still
20.6/23.5\%, and with thirteen phases on Qwen2.5-VL nearly half of all answers
(49.3\%) change somewhere.

\noindent\textbf{Larger budgets do not remove it.} A larger budget might be
expected to wash the phase out. It does not, at least up to 32 frames. As the
budget grows eightfold from 4 to 32 frames (Qwen2.5-VL), pairwise disagreement
falls only from 24.0\% to 12.4\% on EgoSchema and from 34.7\% to 22.1\% on
Video-MME, so each doubling leaves 0.80 to 0.86 of it. At 32 frames, the top of
deployed practice, it is still $3.4\times$ and $6.0\times$ the re-run floor.

\begin{figure*}[t]
\centering
\includegraphics[width=\textwidth]{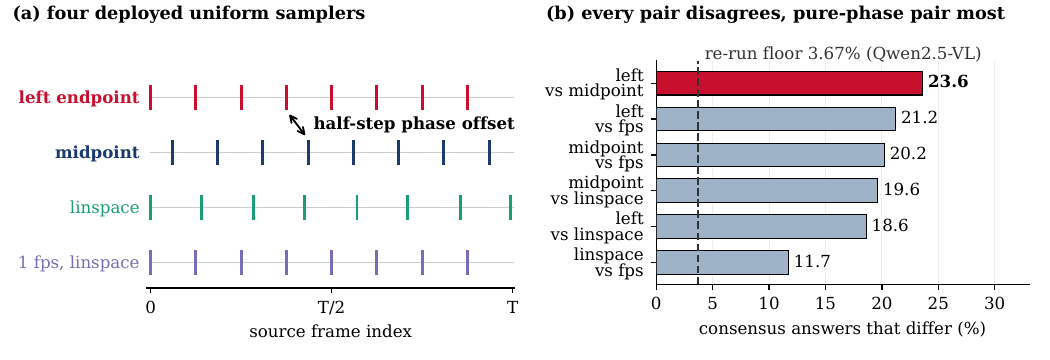}
\caption{\textbf{The disagreement needs no perturbation from us.} (a) Four
uniform samplers taken from deployed harnesses, on a common time axis: left
endpoint $\lfloor iT/b\rfloor$, midpoint $\lfloor (i{+}\tfrac12)T/b\rfloor$,
endpoint-inclusive $\mathrm{round}(i\,\tfrac{T-1}{b-1})$, and decimation to
1\,fps followed by \texttt{linspace}. \emph{Left endpoint} and \emph{midpoint}
have the same rate and differ only by a half-step phase offset. (b) All six pairs disagree far above the re-run floor,
and the pure-phase pair disagrees \emph{most}, on 23.6\% of consensus answers, while
all four score within 1.2 points of one another.}
\label{fig:samplers}
\end{figure*}

\subsection{Samplers in the wild}
The phase already varies between deployed harnesses \cite{khatri2026select}. On
Qwen2.5-VL, the four deployed ``uniform'' samplers of Fig.~\ref{fig:samplers}
disagree on 12--24\% of consensus answers pairwise while their scores stay
within 1.2 points of one another. The most telling pair is left endpoint
against midpoint. The two sample at the same rate and differ only by a
half-step phase, yet they disagree most, on 23.6\% of consensus answers, at
0.7 points of score.

\noindent\textbf{Other re-readings of the clip.}
\label{sec:control}
Any legal re-reading of the clip moves answers about as much as the phase does.
Holding the phase at 0 and drawing two other readings under the same ten
orderings (Qwen2.5-VL), two independent uniform-random subsets disagree on
24.2\% of consensus answers and two grids jittered by one frame per cell on
19.0\%, against 20.1\% between two phases. The phase differs from these in two
ways. Deployed samplers fix it silently, and a polyphase average removes the choice
by construction. The consensus itself is not perfectly reliable, since two
disjoint five-ordering halves at one phase already disagree on 18.5 to 19.4\%
across the three core models, but a phase change adds 6 to 9 points on top.

\noindent\textbf{Standardization versus marginalization.}
Fixing the phase makes a number reproducible without making it less arbitrary.
Depending only on which phase is used, the measured Qwen3-VL lead over
Qwen2.5-VL is 3.97, 6.08 or 6.74 points, a 2.77-point swing (order-averaged,
$n{=}599$ items) from one unreported parameter. Marginalization removes the free
parameter from the protocol, whereas standardization merely picks one of its
values.

\section{Polyphase Marginalization}
\label{sec:method}

\subsection{The offset grids are polyphase components}
\noindent Sample $i$ of the
phase-$s$ grid sits at
\begin{equation}
\Big(i+\tfrac{s}{S}\Big)\frac{T}{b} \;=\; \big(iS+s\big)\frac{T}{Sb},
\label{eq:poly}
\end{equation}
which is sample $k{=}iS{+}s$ of the dense $Sb$-frame uniform grid, and
$k=iS+s$ is a bijection onto $[0,Sb)$. The $S$ offset grids are therefore
exactly the polyphase components of that dense grid, disjoint whenever the clip
has at least $Sb$ frames (up to one frame of integer rounding, Fig.~\ref{fig:poly}). The textbook
remedy for a phase-sensitive decimator is to filter before decimating,
averaging each sample's neighbourhood. A vision encoder consumes discrete frames, not temporally blurred ones, so we average after the encoder instead,
over the posteriors of the $S$ components. The decomposition itself is the
standard polyphase identity of multirate signal processing \cite{crochiere1983multirate, vaidyanathan1993multirate}. The identity fixes the set of views deterministically. How their posteriors should be combined is an empirical question that Section~\ref{sec:exp} answers.

\subsection{\phasefusion{}}
Let a question have $n$ options presented under permutation $\pi$ (option at slot $j$ is content $\pi(j)$), and let the model see frames $F_s$ sampled at grid phase $s/S$, budget $b$: $F_s = \{\,f_{\lfloor (i+s/S)\,T/b \rfloor}\,\}_{i=0}^{b-1}$. One forward pass yields letter log-probabilities at the answer position. Renormalized over the $n$ live letters and mapped through $\pi^{-1}$, they define a content-space distribution $q_s(\cdot \mid \pi)$. \phasefusion{} outputs \begin{equation} \hat{y} \;=\; \arg\max_{c}\; \frac{1}{S}\sum_{s=0}^{S-1} q_s(c \mid \pi), \qquad S{=}3,\; b{=}8, \label{eq:fusion} \end{equation} the mixture over phases (arithmetic pooling, geometric pooling performs within 0.3 points). All $S$ phases enter symmetrically, so the output no longer depends on which of the prespecified phases an evaluator would have selected. The continuous phase integral $\int_0^1 q_\delta(c)\,d\delta$, with phase treated as periodic modulo one, is origin-invariant up to one boundary frame in $b$. Eq.~\ref{eq:fusion} is its finite Riemann approximation and retains a residual origin sensitivity that we measure in Section~\ref{sec:exp}.

\noindent\textbf{Order is untouched.} Pooling phases leaves order sensitivity unchanged (order flips 53.1${\to}$53.4\%, $p{=}1.0$, $n{=}446$), which is why the ordering axis needs its own instrument (Section~\ref{sec:flag}).

\noindent\textbf{Cost.} Attention units are in Table~\ref{tab:main}. End to end, the three passes cost $3\times$ one 8-frame pass. Median serial latency is 3.18\,s, against 0.99\,s for one 8-frame pass, 2.86\,s for the 24-frame pass (10\% less, at 13.3\% rather than 10.1\% origin sensitivity) and 3.92\,s for 32 frames.

\section{Experiments}
\label{sec:exp}

\begin{figure}[t]
\centering
\includegraphics[width=\columnwidth]{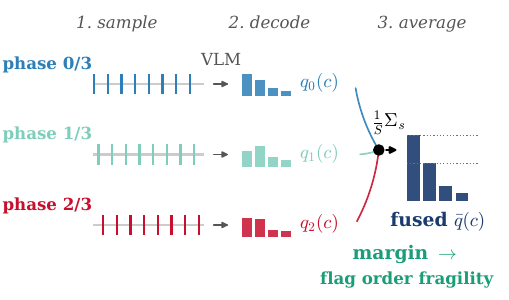}\\[2pt]
\includegraphics[width=\columnwidth]{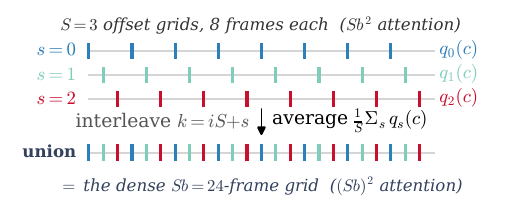}
\caption{\textbf{\phasefusion{}.} Top: the $S$ phase grids are decoded
separately and their content-space posteriors averaged. The margin of the fused posterior is available for flagging order fragility at no extra pass (Section~\ref{sec:flag}). Bottom: the three offset grids interleave into the
dense $Sb$-frame grid, sample $i$ of phase $s$ being sample $k{=}iS{+}s$
(Eq.~\ref{eq:poly}).}
\label{fig:poly}
\end{figure}

\begin{table}[t]
\centering\small\renewcommand{\arraystretch}{1.2}
\setlength{\tabcolsep}{3pt}
\resizebox{\columnwidth}{!}{%
\begin{tabular}{lrrrr}
\toprule
\multicolumn{5}{l}{\textit{A. Cost and accuracy}, Qwen3-VL, $n{=}442$ paired items} \\
arm & attn. & acc.\ \% & $\Delta$ vs.\ 32f & $p$ \\ \midrule
\rowcolor{fuse} \phasefusion{} $3{\times}4$f & 48 & 58.8 & -1.13 & .603 \\
single pass, 8f & 64 & 55.0 & -4.98 & .004 \\
\rowcolor{fuse} \phasefusion{} $3{\times}8$f & 192 & \textbf{60.4} & +0.45 & .883 \\
single pass, 16f & 256 & 57.5 & -2.49 & .109 \\
single pass, 32f & 1024 & 60.0 & +0.00 & -- \\
\midrule
\multicolumn{5}{l}{\textit{B. Reproducibility under a half-step origin shift}, $n{=}587$} \\
arm & & changed \% & & $p$ vs.\ fusion \\ \midrule
single pass, 8f & & 18.2 & & $<$.001 \\
dense pass, 24f (union) & & 13.3 & & .032 \\
\rowcolor{fuse} \phasefusion{} $3{\times}8$f & & \textbf{10.1} & & -- \\
\bottomrule
\end{tabular}}
\caption{\textbf{What averaging buys, on one table.} (A) At $S{=}3$, $b{=}8$ fusion sits on the 32-frame accuracy at $3/16$ of its visual-prefix attention. (B) Under a shift of all three grids by half their spacing, seeing the same 24 frames as one sequence buys the accuracy but not the stability: the dense pass still changes 13.3\% of its answers, fusion 10.1\%. Attention units are $S b^2$ for the visual prefix, and end-to-end latency is reported in Section~\ref{sec:method}. $p$: McNemar, paired. Shaded rows are \phasefusion{}. All blocks are Qwen3-VL at the identity ordering, and each uses the items that carry all of its arms, hence the differing $n$ and accuracies.}
\label{tab:main}
\end{table}

\subsection{Averaging costs no accuracy}
Three short passes match one long pass in accuracy. \phasefusion{} at
$3{\times}8$ frames sits on the accuracy of the 32-frame pass, the largest
single pass we decode, at $3/16$ of its visual-prefix attention, and is
equivalent to it within a prespecified margin (Table~\ref{tab:main}A). Over 500
questions on each of three benchmarks per model, at the identity ordering,
fusion reaches 60.2\% for Qwen3-VL and 54.9\% for Qwen2.5-VL.

The accuracy does not come from extra compute. At 192 attention units fusion
outscores the 16-frame single pass at 256 units by 2.9 points
(Table~\ref{tab:main}A), and it beats a majority vote over the same three views,
59.3\% against 57.8\% on the same 1{,}132 passes, so how the views are pooled matters. Even three 4-frame passes, cheaper than a single 8-frame pass, stay within 1.2 points of the 32-frame pass (Table~\ref{tab:main}A). Returns saturate by $S{=}3$: two more passes at $S{=}5$ add nothing
(60.0 vs.\ 60.4\% on the 442 paired items). The gain also survives a change of
ordering. Under all ten orderings (Qwen2.5-VL, $n{=}445$), per-ordering
accuracy is 51.2\% single-phase against 52.7\% fused, and the ordering flip
rate is unchanged (59.8 vs.\ 58.9\%).

\subsection{Averaging buys reproducibility}
What averaging buys over one long pass is stability. The three grids together
are just the 24-frame grid, so we hold the frame set fixed and compare fusion
with one pass over the \emph{same} 24 frames. The two are equivalent in
logit-scored accuracy, but they part under a shift of origin. When all three grids move together by half their spacing ($n{=}587$), the dense pass changes 13.3\% of its answers
and fusion \textbf{10.1\%}, against 18.2\% for a single 8-frame pass
(Table~\ref{tab:main}B). The same holds when the frames are redrawn rather than
shifted. Across two independent random draws ($n{=}598$), a single 8-frame pass
changes 19.6\% of its answers and fusion over a random 24-frame draw 12.0\%,
with the largest drop on Video-MME (27.0${\to}$15.5\%). How the 24 frames are
split matters little. Random and contiguous partitions at the same cost are no
more stable (10.6\% and 11.2\% against 9.7\% for the polyphase split,
$n{=}570$, within noise), and only the polyphase split adds no sampling seed of
its own.

\subsection{Flagging order fragility in one pass}
\label{sec:flag}
Order cannot be averaged away at this price, since exhaustive marginalization
costs $n!$ passes, but the pass already decoded can tell which answers would
move. The answer margin of a single plain pass ranks which questions flip under
re-ordering at 0.831 AUROC, better than it predicts correctness (0.695), and
ahead of top-1 probability and negative entropy (0.82/0.80, pooled,
$n{=}1{,}350$). It transfers zero-shot to the two families that entered no
design decision, InternVL2.5 and MiMo-VL (0.867/0.876). In practice, flagging
the lowest-margin fifth of items marks a set that flips 89.6\% of the time
against 48.5\% outside it, catching 31.6\% of all flips. Under \phasefusion{}
the same statistic is read off the fused posterior it already computes
(Fig.~\ref{fig:poly}), so in either setting flagging adds no pass.

\subsection{When it helps}
Fusion helps most where adjacent phases lie far apart in time. It can only
change an answer where the phases disagree, which happens on 18--41\% of items,
and the offset between adjacent phases, $\text{duration}/(Sb)$ seconds, is known
from metadata. An exploratory 1\,s split separates the regimes, with a gain of
$-0.3$ points below it ($n{=}370$) and $+2.7$ points above it ($n{=}1{,}427$).
Accordingly, the only significant gains are on long Video-MME clips, $+4.4$ and
$+5.0$ points for Qwen3-VL and MiMo-VL. Qwen2.5-VL gains $+0.6$ there, and the
gains on EgoSchema ($+2.2/{+}1.8$) and short-clip MVBench ($-0.2/{+}1.6$) are
small and not significant. The two halves of the paper meet here. The phase moves the
most answers on long clips at small budgets (Section~\ref{sec:audit}), and that is also where
averaging it pays.

\noindent\textbf{Statistical protocol.} The equivalence claims rest on two
one-sided tests~\cite{schuirmann1987tost} against a margin of $\pm2.5$ points,
fixed \emph{ex ante} at half the 8${\to}$32-frame gain of the cost sweep
($+5.0$ points, Table~\ref{tab:main}A) before any fusion arm was decoded.
Against the 32-frame pass the test pools the two models ($n{=}1{,}132$ passes
over 581 questions, question-clustered, $p{=}.022$, 90\% CI $[-2.09,
+1.56]$). Against the 24-frame union, scored from answer-letter
log-probabilities (the logit path), it passes per model ($p{=}.035/.046$,
$n{=}592/596$, 90\% CIs $[-2.30, +1.62]$ and $[-2.44, +2.44]$). Scored instead
from the generated letter (the generation path), the point estimates are
$+1.12/{-}1.15$ points ($n{=}448/435$) and the margin is not certifiable at
these $n$ ($p{=}.15/.22$). Paired answer changes use McNemar tests (vote
baseline $p{=}.02$, origin shift $p{=}.032$, redraw $p{<}.001$), and the
Video-MME gains reach $p{\leq}.004$. The eight confirmatory tests form one
Benjamini--Hochberg family \cite{benjamini1995controlling}: the pooled
equivalence test (.022), the origin-shift McNemar (.032), the phase lever of
Section~\ref{sec:control} (permutation test, .048), $S{=}3$ against $S{=}2$
(.016) and against its own single views (.022), the MiMo-VL fusion gain (.037),
and the margin against top-1 probability (.005) and against a ten-pass vote
(.042). The largest sits below its step-up threshold of .05, so all eight
survive at $q{<}.05$.

\section{Conclusion}
\label{sec:concl}

A uniform sampler is specified by a rate and a phase. Benchmarks report the
rate and leave the phase to an implementation detail, and that detail changes
roughly one answer in five while leaving the score where it was, most
sharply between two deployed samplers that differ by nothing else. The fix is to report the sampler's phase convention with its budget, or to marginalize it. Because the $S$ offset grids are the
polyphase components of the dense grid, averaging their posteriors damps the parameter at no cost in accuracy inside a prespecified margin. Reporting the convention costs a line, and marginalizing it costs three short passes whose union is the dense grid. Our evidence covers five open 7--8B releases from three families and requires option log-probabilities. Passes and audit will be released upon publication. Audio-language evaluation already shows the option-order
half of this picture~\cite{lin2026hearingtheorder}, and we conjecture the frame-offset half holds there too.

\newpage
\section{Compliance with Ethical Standards}
This is a computational study of publicly released models and benchmarks
(EgoSchema, MVBench and Video-MME), used under their published terms. No new
data were collected from human or animal subjects, and no ethical approval was
required. The authors declare no conflict of interest.

\bibliographystyle{IEEEbib}
\bibliography{refs}

\bigskip
\appendix\setlength{\emergencystretch}{3em}
\renewcommand\thetable{S\arabic{table}}\setcounter{table}{0}
\renewcommand\thefigure{S\arabic{figure}}\setcounter{figure}{0}
\section*{Supplementary Material}
This supplement gives the per-cell breakdowns behind the pooled numbers in the main paper, the full protocol, the intervention ladder that motivates flagging rather than repairing order sensitivity, and the auxiliary analyses the main paper refers to. Section, table, figure and equation numbers without an S prefix refer to the main paper. An executable audit that checks every load-bearing number against the result files will be released with the corpus.

\section{Protocol and Dataset Statistics}
\label{sup:protocol}

\paragraph{Benchmarks and sampling.} Table~\ref{tab:datastats} gives the
composition of each evaluated subset. EgoSchema's 500 items are its complete
public subset; MVBench and Video-MME are uniform random samples drawn with
\texttt{random.Random(42).sample}. The option-count column matters for
interpreting flip rates: 369 MVBench items carry three options, which admit
only $3!{=}6$ distinct permutations, so those items' ten orderings necessarily
contain repeats and their measured flip rate is a lower bound relative to the
4- and 5-option benchmarks. Deduplicating changes nothing (2{,}952/2{,}952
redundant variants agree), and the constrained stratum in fact flips
\emph{most}: 63.5\% at three options ($n{=}1107$) against 50.2\% at four
($n{=}1893$) and 51.7\% at five ($n{=}1500$); dropping every three-option
item still leaves 50.9\% $[49.2, 52.5]$.

\paragraph{How many orderings are enough?} Three checks show the ten-draw
flip rate is a converged property, not a sampling accident. First, on the
three-option stratum the ten draws cover \emph{all} $3!{=}6$ permutations
for every one of its 1{,}107 cells, so the measured 63.5\% \emph{is} the
exhaustive rate, exactly. Second, subsampling $k$ of the ten orderings
(50 seeds) gives 25.3/36.4/42.4/48.8/52.0/54.0\% at
$k{=}2/3/4/6/8/10$: the gain per added ordering shrinks monotonically
($11.1/6.0/3.2/1.6/1.0$ points), and the last two orderings buy only $+2.0$ points. Third, a question-clustered bootstrap puts the pooled nine-cell
rate at 54.0\% $[52.1, 55.8]$ (Qwen corpus 53.0\% $[50.8, 55.0]$), so the
headline is insensitive to the resampling unit
(\path{results/derived/ordering_convergence.json}). At four and five
options the ten-draw rate remains a lower bound on the full-permutation
rate, which only strengthens the reported instability.

\begin{table}[t]
\centering\small
\begin{tabular}{@{}lrrlr@{}}
\toprule
Benchmark & pool & used & options & perms \\
\midrule
EgoSchema & 500   & 500 & 5 (all)      & 10 \\
MVBench   & 1420  & 500 & 3 / 4        & 6 \\
          &       &     & ($369$ / $131$) &   \\
Video-MME & 2700  & 500 & 4 (all)      & 10 \\
\bottomrule
\end{tabular}
\caption{Evaluated subsets. ``perms'' is the number of distinct option
orderings available given the smallest option count present, capped at the ten
we evaluate.}
\label{tab:datastats}
\end{table}

\paragraph{Inference.} Frames are sampled uniformly over the clip with OpenCV
at grid positions $\lfloor (i + s/S)\,T/b \rfloor$ for $i{=}0..b{-}1$ ($T$ the
frame count, $s$ the phase index, $S$ the phase count; $s{=}0$, $S{=}1$
recovers the standard sampler), each resized so the longer side is at most
360px. Frames and the question with its permuted options go through the
model's own chat template; the instruction requests the option letter only.
Decoding is greedy (\texttt{do\_sample=False}, at most 4 new tokens) in
bfloat16 on single 24--48GB GPUs. Orderings are the identity plus nine
permutations from a generator seeded by (question id, ordering index),
identical across models, so all cross-model comparisons are paired at the
(question, ordering) level.

\paragraph{Cloze scoring.} Row 3 of Table~\ref{tab:negatives} scores each option's text with no lettered
list, under three normalizations: raw $\sum_t \log P(o_t \mid v,q,o_{<t})$, its
token mean, and domain-conditional PMI $\log P(o \mid v,q) - \log P(o \mid q)$
\cite{holtzman2021surface}. On 100 Video-MME items (Qwen3-VL-8B) they reach
35.0/37.0/42.0\% against 53.0\% for the letter channel on the same items, so the $-11$ points of Table~\ref{tab:negatives}, row 3, is the \emph{best} of the three, not an artifact of unnormalized scoring. PMI costs two passes per option, one of them text-only.

\paragraph{Reasoning-model prefill.} Any logit-level screen on a
reasoning-tuned model needs the following protocol step. MiMo-VL-7B-RL opens every
response with a \texttt{<think>} scaffold, so with a four-token budget the
model never reaches an answer letter and the first-token log-probs sit
${\approx}30$ nats below the top token, carrying no answer information: every
\texttt{pred\_letter} in an unmodified run is \texttt{"<"}. Under our
answer-only instruction the scaffold the model emits is \emph{empty}, so we
prefill that closed empty block into the assistant turn. On a six-question
probe this made the answer letter the first generated token in 6/6 cases
(0/6 without); over the full run, zero predictions are non-letters and the
logit argmax agrees with the emitted letter on 100\% of passes. The prefill reproduces behaviour the model exhibits unprompted rather than suppressing reasoning it would otherwise perform.

\paragraph{Population sizes.} The core ordering sweep covers three models
by three benchmarks at 500 questions per cell; the separate MiMo-VL logit
replication covers 150 questions per benchmark. In the frame-phase sweep,
$n{=}598$ (Qwen3-VL), $n{=}599$ (Qwen2.5-VL), $n{=}593$ (InternVL2.5), and
$n{=}599$ (InternVL3.5) items carry all three phases, roughly 200 per
benchmark. Per-model phase numbers in the main paper are computed on these
populations, and the 22.2--28.5\% any-of-three range in the main paper spans all four.

\paragraph{A dense random-phase Monte Carlo.} The three-phase grid is
typical of the continuous phase space, not a special construction. Rerunning Qwen2.5-VL on 592 of the same items and their orderings at 13 phases (the zero anchor plus 12 stratified-random draws, seed 11), the pairwise
consensus-disagreement rate is 20.4\%, against 20.1\% on the
$\{0,\tfrac13,\tfrac23\}$ grid; a random 3-subset of the 13 phases gives an anywhere-rate of 29.1\% (90\% interval $[26.4, 31.6]$) over 300 subsets, bracketing the
printed 28.5\%. Coverage grows with density: across all 13 phases,
\textbf{49.3\%} of consensus answers change somewhere, so the three-phase
figures in the main paper are conservative lower bounds
(\path{results/derived/phase_mc.json}).

\paragraph{Implementations in the wild already disagree.} The phase sweep
constructs its perturbation; deployed harnesses need no construction. We
re-decoded the same items and ten orderings under four uniform-sampler
conventions in production use: left-endpoint $\lfloor iT/b\rfloor$ (torchvision-style), midpoint $(i{+}\tfrac12)T/b$ (TSN-style), endpoint-inclusive \texttt{linspace}, and 1-fps decimation followed by \texttt{linspace}. Each is a defensible reading of ``uniformly sample 8 frames.'' On Qwen2.5-VL ($n{=}598$), any two conventions
disagree on 11.7--23.6\% of consensus answers (re-run floor 3.67\%), and
33.3\% of items change consensus under at least one pair, while the four
conventions' accuracies sit within 1.2 points of one another (51.1--52.3\%).
Two harnesses can thus reproduce each other's scores to within a point yet
disagree on one question in five, and every one of them truthfully reports
``uniform sampling'' (\path{results/derived/convention_audit.json}).

\paragraph{A matched-vote control.} The re-run floor quoted throughout is a
reproducibility floor: how far the consensus moves when nothing changes.
Because the ten orderings are held fixed across phases, a stricter null is a
fresh ordering draw at the same vote size, which perturbs the consensus
without carrying any phase information. Splitting each item's ten orderings into
disjoint five-ordering halves (50 random splits per item, seed 0),
consensus disagreement between two halves \emph{within} one phase is
19.4/18.5/19.3\% (Q3/Q2.5/IV2.5), and between halves drawn at
\emph{different} phases it rises to 26.1/27.9/25.1\%. Both halves vote at
size five, so these rates index a noisier statistic than the ten-ordering
consensus and are not comparable to the main paper's phase rates; the
within- versus across-phase contrast is. The phase adds
disagreement over and above the ordering lottery, in every family
(\path{results/derived/matched_vote_control.json}).

\paragraph{The phase lever, in full.} The 3.97/6.08/6.74-point gaps in the main paper are measured on the balanced $n{=}599$ items (the lever
averages accuracy over orderings, so it needs no consensus; the $n{=}598$
in Population sizes counts items with a defined ten-ordering consensus at
every phase), and the
test statistic is the lever itself, $\max{-}\min$ over the three per-phase
gaps (2.77 points unrounded), referred to 20{,}000 within-item permutations of
the phase labels ($p{=}.048$). Measuring the same lever through a
\emph{single} option ordering is itself an ordering lottery: across the
ten draws it ranges 0.3--5.0 points (median 3.4, sd 1.4), is significant in one
of ten, and the order-averaged estimate sits at the 40th percentile of the
ten, so marginalizing the ordering axis is a precondition for measuring the
phase axis. In the matched pairwise unit, the phase moves the consensus on
18.7\%/20.1\% of item pairs (Q3/Q2.5) against the 2.34\%/3.67\% re-run floor,
$8.0\times$/$5.5\times$ the pipeline's own irreproducibility. (\path{results/derived/phase_lever_estimators.json}).

\paragraph{Two more families, and a generation of architecture.}
InternVL2.5-8B replicates the finding outside the Qwen line. Replaying its
released ten orderings at the same three grid phases changes the
ten-ordering consensus answer on \textbf{22.9\%} $[19.7, 26.5]$ of
$n{=}593$ questions, with per-phase accuracy flat (50.7/51.4/51.4\%). As in
the Qwen sweeps, phase 0 is re-run rather than copied, so agreement with the
released ordering corpus is measured rather than assumed: it differs on
0.0\% of 5,930 passes, a floor the phase effect clears outright.

Its successor answers the obvious question, whether a newer generation
fixes this. \textbf{InternVL3.5-8B}, which replaces the InternLM2 language
tower with Qwen3, was run on the same items and the same permutations:
the consensus changes on \textbf{22.2\%} $[19.1, 25.7]$ of $n{=}599$
questions (16.5/19.6/30.5\% by benchmark), per-phase accuracy 52.9/53.4/53.0\% (a 0.49-point spread, the flattest of the four phase populations), and its ordering-axis flip rate at phase 0 is 50.9\%. A generation of architecture and training moves phase stability by 0.7 points, within overlapping intervals, and leaves the ordering axis where it was (the two InternVL phase-replication JSONs in
\path{results/derived/}).

\paragraph{Compute.} The corpus comprises the ordering sweep (nine cells $\times$ 500 questions $\times$ ten orderings), the logit subsets, the frame-phase sweeps for four releases, the phase-by-budget sweep (47{,}930 passes), the 13-phase Monte Carlo, the deployed-sampler audit, the \phasefusion{} captures and $(S,b)$ ablation (including the MiMo-VL fusion arm), and the union, origin, re-reading, partition and timing controls (Section~\ref{sup:union}), 289k forward passes in all.

\paragraph{Multiplicity.} The eight confirmatory tests of the main paper survive Benjamini--Hochberg control \cite{benjamini1995controlling} as a
family at $q{=}.05$:
the largest $p$, $.048$ at rank $8$, falls below its critical value
$q\cdot 8/8{=}.050$, so the step-up procedure rejects all eight
(Table~\ref{tab:bh}; \path{results/derived/bh_family.json}). The family comprises the eight tests the paper's claims rest on, listed in Table~\ref{tab:bh} and enumerated row by row in \texttt{bh\_family.json}; remaining $p$-values are descriptive and reported unadjusted.

\begin{table}[t]
\centering\small
\begin{tabular}{@{}lrr@{}}
\toprule
Confirmatory test & $p$ & crit. \\
\midrule
Margin vs.\ MSP, flip prediction        & .005 & .0063 \\
$S{=}3$ vs.\ $S{=}2$ phase count        & .016 & .0125 \\
Fusion vs.\ 32f, clustered TOST         & .022 & .0188 \\
$S{=}3$ vs.\ its own single views       & .022 & .0250 \\
Fusion vs.\ union pass, origin shift    & .032 & .0312 \\
MiMo-VL fusion gain                     & .037 & .0375 \\
One-pass margin vs.\ ten-pass vote      & .042 & .0438 \\
Phase lever, order-averaged             & .048 & .0500 \\
\bottomrule
\end{tabular}
\caption{Benjamini--Hochberg over the confirmatory family ($m{=}8$,
$q{=}.05$). The largest passing rank is $8$, so the step-up procedure
rejects all eight; intermediate ranks need not clear their own thresholds.}
\label{tab:bh}
\end{table}

\section{Interventions That Do Not Remove Order Sensitivity}
\label{sup:ladder}
Table~\ref{tab:negatives} summarizes thirteen interventions spanning every
level a practitioner controls. Three results determine the method design.
Here \spr{} denotes the single-pass answer margin of Section~\ref{sec:flag} of the main paper.

\paragraph{Aggregation cannot remove presentation noise.} A ten-ordering
vote gains $+2.08$ points over the pooled Qwen corpus ($+1.74$ over all nine
cells), but two disjoint four-ordering votes still disagree on 18.2\% of
items. Even an oracle allocator that knows which questions flip has no
headroom over the best uniform budget ($-0.4$ points, bootstrap CI
$[-1.9, +0.7]$ points). Margin-guided reallocation and routing fail likewise
(rows 4, 9, 10): the margin ranks fragility but is too blunt to buy
accuracy back by re-spending the budget. Repetition without new evidence
leaves a residual ordering choice.

\paragraph{The effect is not explained by answer-letter priors.} Removing labels entirely, presenting an unlabeled list and matching the generated text, reduces flips only from 53.0\% to 49.0\% ($n{=}100$). The model tracks
content more strongly than letters (content versus letter fixation, 0.81
versus 0.46), but remains sensitive to the presentation slots themselves.

\paragraph{Naive consistency tuning can collapse uncertainty.} We train
seven LoRA \cite{hu2022lora} arms from two objective families on
4{,}996 NExT-QA \cite{xiao2021nextqa} MCQs,
disjoint from all three evaluation benchmarks, and score them on the
Video-MME logit cell ($n{=}150$). Distribution matching has a degenerate
flattening optimum: at $\lambda{=}10$, flips rise from 53.3\% to
\textbf{80.0\%} while the median margin falls from 0.83 to 0.60
($p{=}10^{-7}$). Hard pseudo-label consistency also worsens flips at high
confidence gates ($+10.7$ points, $p{=}.024$); its only reduction is not
significant and costs accuracy. (Concurrent RL training with joint
answer-consistency rewards reports gains
\cite{huang2025acre,zheng2026pagrpo}; ours cover frozen-backbone LoRA
distillation.)

Additional ablations point to visual evidence rather than letter bias.
Removing the video raises the flip rate from 50.3\% to \textbf{79.9\%} while
accuracy falls from 57.7\% to 38.5\% ($n{=}1500$, Qwen3-VL). Thus the
instability studied here is the residual regime after visual evidence has
already removed roughly thirty points of blind-text instability.

\begin{table*}[t]
\centering\footnotesize
\setlength{\tabcolsep}{4pt}
\begin{tabular}{@{}cp{4.6cm}lp{6.4cm}p{2.6cm}@{}}
\toprule
\# & Intervention (cost in passes) & Level & Outcome on the failure it targets & Verdict \\
\midrule
1 & $k$-ordering plurality vote (10$\times$) & inference &
    $+2.08$ points Qwen-pooled ($+1.74$ points over all 9 cells); two disjoint 4-votes disagree on 18.2\% of items & lottery persists \\
2 & SPR-gated adaptive re-ask (2--5$\times$) & inference &
    \textbf{oracle} gate 55.0\% at its best budget vs.\ best uniform
    55.4\% ($-0.4$ points) & no headroom \\
3 & Per-option content scoring (cloze, $2n_{\mathrm{opt}}{\times}$) & inference &
    raw 35.0/token-mean 37.0/PMI \textbf{42.0}\% vs.\ letter 53.0\%, same items & $-11$ points at best \\
4 & Margin-routed channel switch ($\sim$1.5$\times$) & inference &
    $+1.9$ points on flipped items, $-25$ points on stable ones & nothing to route to \\
5 & Content-space logit fusion, antithetic pair (2$\times$) & inference &
    54.7 vs.\ 54.5\% (random pair) on exact-reverse subset & pairing worth 0.2 points \\
6 & Phase-ensemble, argmax majority (3$\times$) & inference &
    $3{\times}8$f vote 57.8\% $\approx$ 16f single 58.1\% ($p{=}.79$); logit
    mixture $+1.5$ points over the vote ($p{=}.020$) & counting loses to pooling \\
7 & PriDe letter-prior debiasing (1$\times$ / $k\times$) & inference &
    single-pass $+0.2$ points over naive (56.4 vs.\ 56.2\%); full marginalization
    58.0\% $\approx$ logit-free vote 57.8\% & prior real but immaterial \\
8 & Label-free joint presentation (1$\times$) & prompt &
    flip 53.0\% $\to$ 49.0\%, accuracy unchanged & slot-noise, not letters \\
9 & SPR-driven Neyman estimation (1.5--5$\times$) & measurement &
    RMSE ratio vs.\ uniform 1.15$\times$ at $b{=}1.5$, ${\le}1.02\times$ beyond & margin too blunt \\
10 & SPR-guided flip detection (1.5--3$\times$) & measurement &
    recall ratio ${\le}1.22\times$; deeper escalation \emph{worse} & base rate 57\% \\
11 & Consistency tuning, JS-to-mixture (LoRA, 4 arms) & training &
    $\lambda{=}10$: flip 53.3\%$\to$\textbf{80.0\%} ($p{=}10^{-7}$); median margin 0.83$\to$0.60 & degenerate optimum \\
12 & Consistency tuning, confident pseudo-label (3 arms) & training &
    best arm $-2.6$ points flip (n.s.) at $-2.2$ points accuracy; $\tau{=}0.9$: $+10.7$ points ($p{=}.024$) & consistency $\neq$ correctness \\
13 & Cross-family transfer of certified thresholds (1$\times$) & calibration &
    held-out-family selective error up to 30\% at $\alpha{=}25\%$ & rankings transfer, thresholds don't \\
\bottomrule
\end{tabular}
\caption{\textbf{The repair ladder, tested rung by rung.} Every intervention
level available to a practitioner (inference-time aggregation and
debiasing, prompt re-design, measurement reallocation, training-time
consistency objectives, and calibration transfer) fails to repair
per-question instability, several worsening it. Two
results bound the space we test: an \emph{oracle} re-ask allocator shows
no headroom over the best uniform budget (row 2; bootstrap CI
$[-1.9, +0.7]$ points), and no LoRA consistency objective reduces flips at $p{<}.05$ (rows 11--12), because forcing
agreement at genuine near-ties trades instability for arbitrary commitment. The instability is epistemic signal, not
a removable artifact, which is exactly why it can be \emph{screened}
(Section~\ref{sec:flag}) even though it cannot be \emph{fixed}. \spr{} is the single-pass answer margin of Section~\ref{sec:flag} of the main paper.}
\label{tab:negatives}
\end{table*}

\section{Per-Cell Instability}
\label{sup:percell}

Table~\ref{tab:instability} is the per-cell form of the nine-cell flip rates pooled in Section~\ref{sec:audit} of the main paper. Flip rate is the fraction of questions whose chosen \emph{content
index} (the option identity recovered through the inverse permutation) is not
constant across the ten orderings; swing is the max$-$min accuracy over those
same ten orderings.

\begin{table}[t]
\centering\small
\begin{tabular}{@{}llrrr@{}}
\toprule
Model & Benchmark & Flip\% & Swing (points) & BH $p$ \\
\midrule
Q2.5 & EgoSchema & 52.8 & 5.8 & $<10^{-15}$ \\
Q2.5 & MVBench   & 66.4 & 4.6 & $<10^{-15}$ \\
Q2.5 & Video-MME & 47.8 & 2.6 & $<10^{-15}$ \\
Q3   & EgoSchema & 40.8 & 3.2 & $<10^{-15}$ \\
Q3   & MVBench   & 58.2 & 4.0 & $<10^{-15}$ \\
Q3   & Video-MME & 51.8 & 3.8 & $<10^{-15}$ \\
IV2.5 & EgoSchema & 61.6 & 5.4 & $<10^{-15}$ \\
IV2.5 & MVBench   & 54.8 & 4.0 & $<10^{-15}$ \\
IV2.5 & Video-MME & 51.6 & 2.2 & $<10^{-15}$ \\
\bottomrule
\end{tabular}
\caption{Per-cell flip rate vs.\ accuracy swing. The rightmost column is the
Benjamini--Hochberg-adjusted $p$ for a one-sided binomial test of $H_0$:
flip${}\leq15\%$; all nine fall below $10^{-15}$, and the least extreme
is $3.4\times10^{-44}$ (Q3/EgoSchema).}
\label{tab:instability}
\end{table}

\paragraph{Cross-model decomposition.} Evaluating models on identical items
\emph{and} identical orderings decomposes per-question flip state into
both-stable, shared-unstable (both models flip) and model-specific (exactly
one flips) components (Table~\ref{tab:decomp}). Shared-unstable mass exceeds
the independence null in every pair and benchmark (lift ${>}1$, all bootstrap
CIs excluding 1), so part of the fragility is item-intrinsic; the large
model-specific mass shows it is not reducible to item weakness. The pattern
holds across architectures, not only within the Qwen family.

\begin{table}[t]
\centering\small
\begin{tabular}{@{}llrrr@{}}
\toprule
Pair & Benchmark & Shared & Model-sp. & Lift \\
\midrule
Q2.5$\times$Q3   & EgoSchema & 32.0 & 29.6 & 1.49 \\
Q2.5$\times$Q3   & MVBench   & 43.2 & 38.2 & 1.12 \\
Q2.5$\times$Q3   & Video-MME & 35.0 & 29.6 & 1.41 \\
IV2.5$\times$Q3  & EgoSchema & 33.0 & 36.4 & 1.31 \\
IV2.5$\times$Q3  & MVBench   & 40.6 & 31.8 & 1.27 \\
IV2.5$\times$Q3  & Video-MME & 36.4 & 30.6 & 1.36 \\
IV2.5$\times$Q2.5& EgoSchema & 41.4 & 31.6 & 1.27 \\
IV2.5$\times$Q2.5& MVBench   & 42.8 & 35.6 & 1.18 \\
IV2.5$\times$Q2.5& Video-MME & 32.0 & 35.4 & 1.30 \\
\bottomrule
\end{tabular}
\caption{Flip-state decomposition (\% of shared items). Lift is the
shared-unstable mass over its independence null.}
\label{tab:decomp}
\end{table}

\paragraph{Oracle-bound uncertainty.} The oracle re-ask bound (Table~\ref{tab:negatives}, row 2) carries a
question-clustered bootstrap: the best oracle-gated budget trails the best
uniform budget by $0.37$ points, 95\% CI $[-1.85, +0.67]$ points
(\path{results/derived/oracle_bound_ci.json}): no headroom.

\paragraph{Consistency-tuning setup.} The consistency-tuning negative results (Table~\ref{tab:negatives}, rows 11--12) rest on
seven LoRA arms: four for the JS-to-mixture objective
($\lambda \in \{0, 1, 3, 10\}$, where $\lambda{=}0$ is the plain-SFT control
that isolates the consistency term) and three for confident pseudo-labelling
($\tau{=}0.7$ at $\lambda{=}1$ and $\lambda{=}2$, and $\tau{=}0.9$ at
$\lambda{=}1$). All seven are trained identically, so the arms differ only
in the objective and its strength. Each adapts Qwen2.5-VL-7B with rank-16 LoRA \cite{hu2022lora}
($\alpha{=}32$, dropout $0.05$) on \emph{all seven} linear projections of
every language block (\texttt{q,k,v,o,gate,up,down}), the vision tower
frozen, at learning rate $10^{-4}$ for one epoch over 4{,}996 NExT-QA \cite{xiao2021nextqa} MCQs
with effective batch 16 and 8 frames per clip, bfloat16 with gradient
checkpointing, seed 0. The configuration is deliberately generous: a
higher-capacity adapter reaching every projection is the setting most likely
to \emph{succeed}, which is what makes the negative result informative.
The trainer (\path{code/train_pct.py}) will be released with the code.

\section{Mechanism: Per-Cell Fixation}
\label{sup:mechanism}

Table~\ref{tab:mechanism} gives the fixation statistics per cell (pooled: content 0.81, letter 0.46, position mass 0.38).
Content-fixation (c-fix) and letter-fixation (l-fix) are the modal vote-shares
over orderings of, respectively, the chosen content index and the emitted
letter. Position-mass fixation is the logit-level analogue: the probability
mass placed on the modal letter slot across orderings, regardless of which
option occupies it. The ordering c-fix ${>}$ l-fix ${>}$ pos-mass holds in
every cell.

\begin{table}[t]
\centering\small
\begin{tabular}{@{}llrrr@{}}
\toprule
Model & Benchmark & c-fix & l-fix & pos-mass \\
\midrule
Q2.5 & EgoSchema & 0.82 & 0.40 & 0.31 \\
Q2.5 & MVBench   & 0.78 & 0.54 & 0.38 \\
Q2.5 & Video-MME & 0.82 & 0.44 & 0.34 \\
Q3   & EgoSchema & 0.86 & 0.40 & 0.38 \\
Q3   & MVBench   & 0.78 & 0.53 & 0.47 \\
Q3   & Video-MME & 0.81 & 0.44 & 0.40 \\
\midrule
\multicolumn{2}{@{}l}{Pooled} & 0.81 & 0.46 & 0.38 \\
\bottomrule
\end{tabular}
\caption{Content-, letter- and position-fixation per cell.}
\label{tab:mechanism}
\end{table}

\paragraph{Letter prior.} The prior is real but secondary. A $\chi^2$ goodness-of-fit
test of predicted-letter uniformity, computed on shuffled orderings only (the
identity ordering is excluded to remove the benchmarks' own gold-letter
distribution as a confound) and pooled per option-count group, rejects
uniformity in all eight option-count groups, with $p$ ranging from
$6.5\times10^{-4}$ (Q2.5/MVBench, 4-option) to $3.0\times10^{-96}$
(Q3/MVBench, 3-option). A letter prior therefore exists and is highly significant. The claim is one of relative magnitude: content fixation exceeds letter fixation in every cell. Consistently, PriDe debiasing (Table~\ref{tab:negatives}, row 7)
removes almost nothing: single-pass PriDe moves pooled accuracy from 56.2\% to
56.4\%, and full permutation marginalization reaches 58.0\%, statistically
indistinguishable from the logit-free ordering ensemble at 57.8\%.

\section{The Union Control}
\label{sup:union}

\begin{figure}[t]
\centering
\includegraphics[width=\columnwidth]{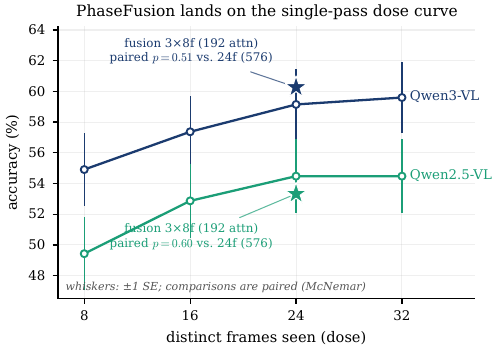}
\caption{\textbf{The union control, visually.} Single-pass accuracy against
the number of distinct frames seen (8/16/24/32; whiskers $\pm1$\,SE), with
\phasefusion{} $3{\times}8$f plotted at its own dose of 24 distinct frames
(star). On the generation-scored path, the star differs from the union pass
by $+1.12/-1.15$ points; McNemar $p{=}.51/.60$ tests difference, not
equivalence. The same-logit TOST ($p{=}.035$/$.046$; ``Same-pipeline
replication'' below) establishes equivalence within the prespecified
$\pm2.5$-point margin. Fusion uses 192 attention units instead of 576 and reduces origin sensitivity.}
\label{fig:union}
\end{figure}

The three phase grids of $S{=}3$, $b{=}8$ partition, up to floor rounding of
at most one frame index, the uniform 24-frame grid. A single pass over those
24 frames is therefore the frame-matched control separating ``fusion pools
complementary evidence'' from ``the model saw more frames''. We ran it on
both families and both scoring paths (Table~\ref{tab:union},
Figure~\ref{fig:union}).

\begin{table}[h]
\centering\small
\setlength{\tabcolsep}{4pt}
\begin{tabular}{@{}lrrrrr@{}}
\toprule
 & 8f & 16f & 24f & 32f & \phasefusion{} $3{\times}8$f \\
\midrule
Q3 ($n{=}448$)   & 54.9 & 57.4 & 59.2 & 59.6 & 60.3 \\
Q2.5 ($n{=}435$) & 49.4 & 52.9 & 54.5 & 54.5 & 53.3 \\
\midrule
Attn.\ units & 64 & 256 & 576 & 1024 & \textbf{192} \\
\bottomrule
\end{tabular}
\caption{Union control, generation-scored pipeline, identity ordering.
Fusion vs.\ 24f: Q3 $+1.12$ points (McNemar $p{=}.51$), Q2.5 $-1.15$ points
($p{=}.60$). Fusion vs.\ 32f: $+0.67$ points ($p{=}.77$) / $-1.15$ points ($p{=}.61$).
Fusion vs.\ 8f: $+5.36$ points ($p{=}4\times10^{-4}$) / $+3.91$ points ($p{=}.043$).
The gain over the single pass is attributable to the additional frame dose,
delivered at $1/3$ the attention.}
\label{tab:union}
\end{table}

\paragraph{Same-pipeline replication.} Scoring the 24-frame pass through the
identical letter-logprob path as fusion (rather than generation) gives
$-0.34$ points, $p{=}.89$ ($n{=}592$, Q3) and $\pm0.00$ points, $p{=}1.0$ ($n{=}596$,
Q2.5); on items shared by both scorers the two judgments agree on
$447/447{=}100\%$, so the near-zero same-path contrast is not a scoring
artifact. This path is also powered against the pre-registered margin: the
paired per-item differences carry standard errors of $1.19$ points ($n{=}592$,
50 discordant pairs) and $1.48$ points ($n{=}596$, 78 discordant), so two
one-sided tests \cite{schuirmann1987tost} reject differences larger than
$\pm2.5$ points at $p{=}.035$ (Q3) and $p{=}.046$ (Q2.5), the equivalence test the main paper cites
(\path{results/derived/b24_crosspipeline.json}). The generation-scored subsets are smaller and do not certify the margin on their own (TOST $p{=}.154$/$.222$), so the equivalence is established on the shared logit path and the generation-path contrast is reported as a point estimate. Per benchmark on the logit path (Q3) the contrast is $-0.51$, $-0.52$ and $+0.00$ points on EgoSchema, MVBench and Video-MME, all $p{=}1.0$.

\paragraph{Timing and memory.} Median seconds per question and peak
memory for every configuration, from one matched idle-node run (100
Video-MME questions, one NVIDIA L40S at batch 1, frame decode, vision
encoder and prompt re-encode all inside every reported second). Medians agree with
means within 2\% for every configuration except $3{\times}8$f, whose
10.6\,s mean reflects a single stalled decode. Serially, three short
passes pay three fixed overheads, so fusion is not faster than the
24-frame union pass it factorizes (3.18 vs.\ 2.86\,s); its cost case is attention compute and the removal of the phase draw.

\begin{table}[h]
\centering\small
\begin{tabular}{@{}lrrr@{}}
\toprule
Configuration & Attn.\ units & Median s/q & Peak GB \\
\midrule
single 4f & 16 & 0.55 & 17.67 \\
single 8f & 64 & 0.99 & 17.83 \\
single 16f & 256 & 1.92 & 18.29 \\
single 24f & 576 & 2.86 & 18.93 \\
single 32f & 1024 & 3.92 & 19.77 \\
fusion 3$\times$4f & 48 & 1.72 & 17.67 \\
fusion 3$\times$8f & 192 & 3.18 & 17.83 \\
\bottomrule
\end{tabular}
\caption{Measured cost of each configuration in the matched run. Means
carry heavy decode tails on long videos, so medians are reported; peak
memory is flat because 8B bfloat16 weights dominate activations at these
lengths.}
\label{tab:timing}
\end{table}

\paragraph{The $(S,b)$ frontier.} Table~\ref{tab:fusion} is the full sweep behind Table~\ref{tab:main}A of the main paper: accuracy against attention cost for every
single-pass budget and every $(S,b)$ fusion configuration we ran, on the
$n{=}442$ subset carrying all of them.

\begin{table}[t]
\centering\small
\setlength{\tabcolsep}{3.5pt}
\begin{tabular}{@{}lrrr@{}}
\toprule
Configuration & Attn. & Acc.\ (\%) & vs.\ 32f \\
\midrule
Single pass, 4f            &   16 & 53.6 & $-6.3$\rlap{$^{*}$} \\
\textbf{\phasefusion, $3{\times}4$f} & \textbf{48} & \textbf{58.8} & $-1.1$ \\
Single pass, 8f            &   64 & 55.0 & $-5.0$\rlap{$^{*}$} \\
Argmax vote, $3{\times}8$f &  192 & 59.3 & $-0.7$ \\
\phasefusion, $2{\times}8$f & 128 & 55.9 & $-4.1$\rlap{$^{*}$} \\
\textbf{\phasefusion, $3{\times}8$f} & \textbf{192} & \textbf{60.4} & $+0.5$ \\
Single pass, 16f           &  256 & 57.5 & $-2.5$ \\
\phasefusion, $5{\times}8$f & 320 & 60.0 & $+0.0$ \\
\phasefusion, $3{\times}16$f & 768 & 60.9 & $+0.9$ \\
Single pass, 32f           & 1024 & 60.0 & --- \\
\bottomrule
\end{tabular}
\caption{\textbf{The $(S,b)$ frontier} (Qwen3-VL, paired $n{=}442$ items
carrying every configuration and baseline; attention cost $S b^2$ in
frame-token$^2$; $^{*}$ below the 32-frame pass at McNemar $p{<}.05$). Every $S{=}3$ configuration sits above the single-pass curve at
its cost. \phasefusion{} at $3{\times}4$f sits 1.1 points below the 32-frame pass and runs at 1.72\,s/question against 3.92\,s in the matched run, and
costs no more than a 16-frame pass (1.92\,s; $+1.4$ points, $p{=}.51$). $S{=}2$ is the exception
(Section~\ref{sup:sfail}); $S{=}5$ buys nothing. Argmax vote is shown at
this subset's own value; its pooled gap is in Section~\ref{sec:exp} of the main paper.}
\label{tab:fusion}
\end{table}

\paragraph{What fusion leaves unchanged, and its per-family gain.}
\phasefusion{} does not touch the ordering axis: order flips move 53.1${\to}$53.4\% ($p{=}1.0$),
an empirical check that pooling phases neither repairs nor worsens
presentation noise. Over the single 8-frame pass it replaces, the
order-averaged gain is $+2.6$ points (Q3) / $+1.5$ points (Q2.5)
(\path{results/derived/phase_fusion_final.json}). That average is not
carried by a lucky ordering: taken one ordering at a time, the gain is
positive under all ten for Qwen3-VL ($+1.35$ to $+4.93$ points, mean $+2.62$)
and under eight of ten for Qwen2.5-VL ($-0.67$ to $+4.49$ points, mean $+1.57$),
a $-0.7$ to $+4.9$-point range
(\path{results/derived/per_ordering_gains.json}).

\paragraph{Origin robustness of the union pass.} Under the matched
half-step origin shift ($T/48$ absolute, the same shift that maximally
perturbs the $S{=}3$ phase set), and paired on the $n{=}587$ items carrying
every arm, the answer changes on 18.2\% of items for a single 8-frame pass,
13.3\% for the 24-frame union pass, and 10.1\% for \phasefusion{}. Both
contrasts against fusion are significant (union pass, McNemar $p{=}.032$;
single pass, $p{=}1.3\times10^{-7}$), and accuracy is flat across origins
(61.18 vs.\ 61.34\% for the union pass). Seeing the same 24 frames as one
sequence therefore buys the accuracy but \emph{not} the reproducibility:
averaging over the phase offsets is what damps the residual origin
dependence, which is the one axis on which the factorization is strictly
better than the pass it factorizes.

\paragraph{Visual distinctness, in full.} Each video is scored with no model
in the loop as the mean $1-$histogram-intersection between corresponding
frames of the $S$ grids. Distinctness tracks the phase offset
($\rho{=}0.77$) and predicts the \emph{rate} of phase disagreement only
weakly ($r{=}0.13$, $p{=}10^{-7}$), but its usability strongly: across
quartiles the disagreement rate barely moves for the lowest two (20.1 vs.\
20.9\%) while the fusion gain goes $-0.2$, $+2.7$, $+2.4$, $+3.6$ points.

\paragraph{The one-second split, in full.} Thresholding the phase offset at
one second partitions the 1{,}797 swept items: below it \phasefusion{} does
nothing ($-0.27$ points, $n{=}370$, $p{=}1.0$), above it it gains $+2.73$ points
($n{=}1427$, $p{=}9\times10^{-4}$). The split is exploratory, and the quartile analysis above gives the threshold-free form of the same effect, with gains ordered $-0.2$, $+2.7$, $+2.4$, $+3.6$ points.

\paragraph{Equivalence, in full.} The paired fusion-vs-32f difference is
$-0.27$ points over 1{,}132 observations on 581 distinct questions, with
question-clustered standard errors giving a 90\% CI of $[-2.09, +1.56]$ points;
two one-sided tests reject differences larger than the pre-registered
$\pm2.5$-point margin at $p{=}.022$. On the union control the shared logit path clears the same margin per model ($p{=}.035$/$.046$, ``Same-pipeline replication'' above).

\section{Origin Robustness of \phasefusion}
\label{sup:origin}

Eq.~\ref{eq:fusion} of the main paper is a three-point Riemann sum for $\int_0^1 q_\delta(c)\,d\delta$, which is origin-invariant up to one boundary frame in $b$; at finite $S$ a residual dependence on the grid
origin $\delta$ survives. We measure it on Qwen3-VL by repeating the full
500-question, three-benchmark capture at $\delta{=}1/6$ (the origin
maximally offset from $\delta{=}0$ for $S{=}3$, hence the worst case) and
comparing fused answers on the $n{=}1488$ shared items.

A single pass changes its answer between the two origins on 17.5\%
$[15.6, 19.5]$ of items; \phasefusion{} changes on 11.0\% $[9.5, 12.7]$, a
$1.6\times$ damping. Accuracy is flat across origins (60.15 vs.\ 60.28\%), so
the residual is seed rather than signal. Three checks confirm the residual is
the expected finite-$S$ discretization rather than an implementation artifact.
(i) The two captures genuinely differ: corresponding frames of the two grids
differ by a mean absolute pixel difference of 50.0/255. (ii) The residual is
concentrated exactly where the phases carry different evidence: origin shifts
change the fused answer on 3.1\% of items whose within-run phases already
agree (close to the 2.34\% byte-identical re-run floor) against 35.2\% of
items where they disagree. (iii) The two grids, $\{0,\tfrac13,\tfrac23\}$ and
$\{\tfrac16,\tfrac12,\tfrac56\}$, union to a uniform $S{=}6$ grid; each $S{=}3$ half deviates from the $S{=}6$ fusion on 5.0\% and 6.2\% of items, about half of their mutual 11.0\% disagreement, a consistency check rather than a test. The $S{=}6$ fusion
scores 60.28\%, differing from $S{=}3$ on 4\% of items ($+0.13$ points,
$p{=}.89$). Origin-invariance is approached as $S$ grows.

\section{Why Three Phases? The Phase-Count Diagnostic}
\label{sup:sfail}

On a matched $n{=}593$ subset (Qwen3-VL, $b{=}8$), the $S{=}2$ mixture over
phases $\{0,\tfrac12\}$ scores 58.35\% against 58.52\% and 59.02\% for its own
two views, with differences of $-0.17$ points ($p{=}1.0$) and $-0.67$ points ($p{=}.64$),
i.e.\ indistinguishable from simply picking one of them. With three phases
$\{0,\tfrac13,\tfrac23\}$ the mixture reaches 61.38\% against views of 58.52,
59.19 and 60.37\%, beating two of the three ($+2.9$ points, $p{=}.022$; $+2.2$ points,
$p{=}.049$) and nominally above the third ($+1.0$ points, $p{=}.42$); head-to-head
it beats the $S{=}2$ mixture by $+3.0$ points ($p{=}.016$), the contrast in the main paper's confirmatory family. On the items where the phases disagree, the $S{=}3$ mixture
scores 47.7\% against 36.2 / 38.9 / 43.6\% for the individual views, while the
$S{=}2$ mixture (33.9\%) is below both of its own (34.7 / 37.3\%). The fused
argmax is never an option no view chose (0 of 593 items in both settings), so
the failure at $S{=}2$ is not a ``compromise candidate'' artifact: with two
views there is simply no majority to form.

\section{Risk--Coverage}
\label{sup:rc}

\begin{figure*}[t] \centering \includegraphics[width=\textwidth]{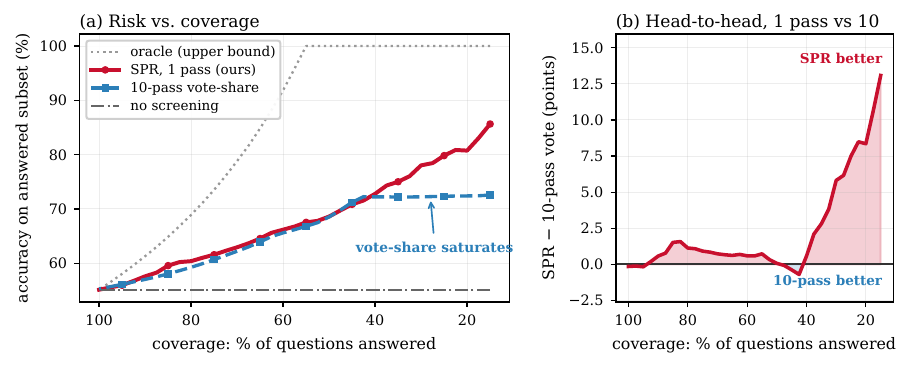}
\caption{Risk--coverage over the pooled three-model logit subset
($n{=}1350$): accuracy on the answered subset as low-confidence items are
abstained on, for the single-pass margin vs.\ the ten-pass vote-share.}
\label{fig:rc}
\end{figure*}

Figure~\ref{fig:rc} compares the single-pass margin against the ten-pass
self-consistency vote-share as selective-prediction signals, each protocol
scored on the answers it would deploy. The vote-share saturates near 72\%
accuracy because it takes only $k{+}1$ distinct values and cannot rank items
inside the unanimous bucket, while the continuous margin keeps ordering them.
This is why the margin pulls ahead by up to 13.1 points exactly where
abstention matters most.

\paragraph{Certified selective risk.} A ranking becomes a guarantee only
after calibration. Within each (model, benchmark) cell we split its 150 logit-scored items in half (75 gold calibration labels, 75 test, 300 random splits) and run fixed-sequence Learn-then-Test: candidate
thresholds are ordered strict to permissive, each is accepted while the
Clopper--Pearson upper bound on calibration selective error stays at or
below $\alpha$, and the walk stops at the first failure. The selected
threshold controls $P(\text{answer}{\neq}\text{gold}\mid\text{answered})$
at level $\alpha$ with confidence $1{-}\delta$, $\delta{=}10\%$
(\path{code/conformal_selective_sim.py}).

\begin{table}[h]
\centering\small
\setlength{\tabcolsep}{4.5pt}
\begin{tabular}{@{}lrrrr@{}}
\toprule
& \multicolumn{2}{c}{$\alpha{=}.30$} & \multicolumn{2}{c}{$\alpha{=}.25$} \\
\cmidrule(lr){2-3}\cmidrule(lr){4-5}
Signal & viol.\ & cov.\ & viol.\ & cov.\ \\
\midrule
\textbf{Margin (\spr{})}  & 6.0 & \textbf{11.6} & 4.7 & 6.7 \\
Top-1 prob.\ (MSP)        & 6.7 & 12.3 & 3.0 & 9.5 \\
Answer entropy            & 6.3 & 13.2 & 5.3 & 9.9 \\
10-pass vote-share        & 1.7 & 5.9  & 1.7 & 2.6 \\
Random                    & 0.0 & 0.1  & 0.0 & 0.1 \\
\bottomrule
\end{tabular}
\caption{In-cell certified mode (\%; 300 splits, 75 calibration labels per
cell). Violations stay inside the $\delta{=}10\%$ budget for every signal.
At matched validity the one-pass margin certifies twice the coverage of the
ten-pass vote-share, at a tenth of the cost. At $\alpha{=}.20$ every signal certifies zero coverage. The fixed-sequence walk starts at the ten highest-scoring calibration items, and the Clopper--Pearson upper bound for ten items with no errors at $\delta{=}10\%$ is $0.206$, above $\alpha$, so the first hypothesis in the sequence fails and the walk stops before any threshold can be accepted. The binding constraint there is the start of the sequence under this $\delta$, not the signal (\texttt{results/derived/\allowbreak conformal\_selective.json}).}
\label{tab:conformal}
\end{table}

\paragraph{Why thresholds are calibrated per cell.} The margin's \emph{ranking} transfers zero-shot across families (Section~\ref{sup:transfer}) while its \emph{scale} does not (row 13 of Table~\ref{tab:negatives}), so thresholds are calibrated per cell.

\section{Fragility-Ranking Transfer}
\label{sup:transfer}

The training-free margin retains its ranking on models untouched by score
selection, including a reasoning/RL model whose elicitation protocol differs.
Table~\ref{tab:ablation} places it among every cheap reliability signal on the
same 1{,}350 questions; Table~\ref{tab:transfer} separately tests whether a
learned four-feature predictor transfers when its training family changes.

\begin{table}[t]
\centering\small
\begin{tabular}{@{}lccc@{}}
\toprule
Signal & Passes & flip & correct \\
\midrule
Random                       & 0  & 0.484 & 0.496 \\
Option count only            & 0  & 0.450 & 0.423 \\
Surface features only$^\dagger$ & 0  & 0.619 & --- \\
\midrule
Answer entropy               & 1  & 0.803 & 0.693 \\
Top-1 prob.\ (MSP)           & 1  & 0.823 & 0.700 \\
Learned, 4 features          & 1  & 0.829 & 0.696 \\
\textbf{Margin (\spr)}       & \textbf{1}  & \textbf{0.831} & 0.695 \\
Fused margin ($3{\times}8$f) & 3  & \textbf{0.854}/0.819 & 0.688/0.691 \\
\midrule
10-pass vote-share           & 10 & ---   & 0.665 \\
\bottomrule
\end{tabular}
\caption{Reliability-signal ablation (AUROC; $n{=}1{,}350$ pooled, three models; fused-margin row per model on its own fused answers, $n{=}446/445$).
No single-pass signal separates on correctness; the margin wins on flip
prediction (vs.\ MSP $p{=}.005$), learning adds nothing ($p{=}.26$), and the
10-pass vote-share loses to one pass on the shared target ($p{=}.042$).
$^\dagger$Qwen-only 5-fold CV ($n{=}3000$); transfers to InternVL2.5 ($n{=}1500$) at 0.551.}
\label{tab:ablation}
\end{table}

\begin{table}[t]
\centering\small
\begin{tabular}{@{}lcc@{}}
\toprule
Transfer setting & Passes & flip AUROC \\
\midrule
in-cell, per cell (range of 9)        & 1 & 0.793--0.925 \\
cross-cell (train 8, test 1; mean)    & 1 & 0.87 \\
cross-model Q2.5$\to$Q3               & 1 & 0.900 \\
cross-model Q3$\to$Q2.5               & 1 & 0.837 \\
\textbf{held-out family} (worst/best of 3) & 1 & \textbf{0.842/0.898} \\
\textbf{held-out RL family} (MiMo-VL) & 1 & \textbf{0.882} \\
\bottomrule
\end{tabular}
\caption{Transfer of the learned four-feature flip predictor when fit on
other cells, models, or entire families ($n{=}450$ per held-out family).
The training-free margin itself needs no fit; its untouched-family results
are 0.867 (InternVL2.5) and 0.876 (MiMo-VL), as reported in Section~\ref{sec:flag} of the main paper.}
\label{tab:transfer}
\end{table}

\section{Per-Cell Fragility Ranking}
\label{sup:spr}

Table~\ref{tab:sprcell} reports the training-free single-pass ranking
in-domain per cell, which the pooled figures understate. Because the margin is not
calibrated across models, pooling nine cells with different margin scales and
base rates costs roughly four AUROC points relative to the mean within-cell
value. The score ranks questions \emph{within} a (model, benchmark) pair; risk control additionally needs the per-cell calibration of Section~\ref{sup:rc}.

\begin{table}[t]
\centering\small
\begin{tabular}{@{}llrrr@{}}
\toprule
Model & Benchmark & flip & correct & base flip\% \\
\midrule
Q2.5 & EgoSchema & 0.847 & 0.723 & 54.7 \\
Q2.5 & MVBench   & 0.793 & 0.688 & 68.7 \\
Q2.5 & Video-MME & 0.875 & 0.659 & 54.0 \\
Q3   & EgoSchema & 0.893 & 0.721 & 41.3 \\
Q3   & MVBench   & 0.877 & 0.599 & 62.0 \\
Q3   & Video-MME & 0.916 & 0.737 & 55.3 \\
IV2.5 & EgoSchema & 0.869 & 0.710 & 64.0 \\
IV2.5 & MVBench   & 0.841 & 0.735 & 58.7 \\
IV2.5 & Video-MME & 0.925 & 0.711 & 52.0 \\
\midrule
\multicolumn{2}{@{}l}{Mean within-cell} & 0.871 & 0.698 & \\
\multicolumn{2}{@{}l}{Pooled (main paper)} & 0.831 & 0.695 & \\
\bottomrule
\end{tabular}
\caption{Single-pass margin in-domain per cell ($n{=}150$ each),
training-free.}
\label{tab:sprcell}
\end{table}

\section{Relation to CircularEval and Fluctuation Rate}
\label{sup:circular}

Our metrics map onto the community's existing instruments as follows.
CircularEval's unanimity rule (correct under \emph{all} circular shifts)
scores 34.0\% $[32.3, 35.8]$ pooled against a 55.99\% naive per-ordering
accuracy (all ten orderings of 3{,}000 questions, i.e.\ the expected
accuracy of an arbitrary single ordering, not the identity-ordering accuracies of the main paper): a 22.0-point gap that measures robustness, not capability, and that
returns no answer at all for the items that flip. The two-ordering
forward/reverse fluctuation rate is 32.2\% on the 1{,}298 items admitting an
exact reverse, while the ten-ordering flip rate on those same items is 58.3\%
for a 26.1-point undercount. A Kendall-$\tau$-maximal two-ordering proxy gives
28.7\%, a 24.3-point undercount on the full set. Cheap two-ordering checks are
lower bounds, not measurements.

\section{\textsc{VidMCQ-Flip} Datasheet}
\label{sup:datasheet}

\paragraph{Composition.} The corpus, to be released upon publication, contains every forward pass behind the paper as paired records: the core ordering sweep (nine model$\times$benchmark
cells at 500 questions $\times$ ten orderings) plus the separate MiMo-VL
logit replication, the frame-budget sweep (budgets 1--32), the frame-phase sweeps (four model releases, three phases, and two origins for Qwen3-VL), the 3-phase $\times$ 10-ordering grid
(150 questions per cell with letter log-probabilities), the
\phasefusion{} and union-control captures (including the 24-frame control
at both origins), the re-reading and partition controls, and the timing runs, 289k passes in all. All files are plain JSON, and \path{results/derived/} holds the derived analysis JSONs behind every table and figure.

\paragraph{Record schema.} Fields vary by capture; every record carries
question id, benchmark, option count, and gold content index, and then the
per-directory fields of Table~\ref{tab:schema}. Benchmark content itself
is referenced by question id under the original licenses; no video or
question text is redistributed.

\begin{table}[h]
\centering\footnotesize
\setlength{\tabcolsep}{3pt}
\begin{tabular}{@{}lp{4.9cm}@{}}
\toprule
Directory & Per-pass fields \\
\midrule
\texttt{extended\_ordering/} & \texttt{ord\_idx}, \texttt{perm},
\texttt{pred\_letter}, \texttt{pred\_content\_idx},
\texttt{new\_gt\_letter}, \texttt{correct} \\
\texttt{logits/} & the above plus \texttt{letter\_logprobs} \\
\texttt{frame\_seed/} & per phase: \texttt{content\_idxs} (ten
orderings), \texttt{mean\_acc} \\
\texttt{phase\_logits/} & per phase: \texttt{pred\_letter},
\texttt{letter\_logprobs} \\
\texttt{frame\_order/} & \texttt{frame\_budget},
\texttt{content\_idxs}, \texttt{flip}, \texttt{mean\_acc} \\
\bottomrule
\end{tabular}
\caption{Per-directory record fields. Permutations are stored where
orderings vary; the phase captures store the phase index instead.}
\label{tab:schema}
\end{table}

\paragraph{Intended uses.} Because every comparison is paired at question
level, the corpus supports flip prediction, confidence calibration,
selective-prediction and conformal research, and evaluation-protocol design
without any GPU inference. An audit script will ship with the corpus: it checks the load-bearing numbers against the result files and, in \texttt{--from-raw} mode, re-derives the headline rates from these records, failing loudly on any mismatch.

\paragraph{Distribution and access.} The corpus and code will be released upon publication as a public Git repository whose tagged releases are archived to a DOI-issuing archive (Zenodo), under a CC-BY-4.0 license for the records and MIT for the code.
Benchmark questions and videos are not redistributed; records reference
them by question id under the benchmarks' original licenses.

\paragraph{Maintenance.} The authors will maintain the repository, version
new captures by date, and preserve the audited JSONs immutably so that the
audit script keeps passing against the released snapshot.

\end{document}